\pdfoutput=1
\documentclass[11pt]{article}

\usepackage{acl}

\usepackage{times}
\usepackage{latexsym}

\usepackage[T1]{fontenc}
\usepackage[utf8]{inputenc}

\usepackage{microtype}

\usepackage{url}
\usepackage{multirow,booktabs,makecell}
\usepackage{booktabs}
\usepackage{graphicx}
\usepackage{adjustbox}
\usepackage{color}
\usepackage{soul}
\usepackage{comment}
\usepackage[labelformat=simple]{subcaption}

\usepackage[T1]{fontenc}
\usepackage{enumitem}
\usepackage[normalem]{ulem}
\usepackage{ amssymb }

\newcommand{\obj}[1]{\textcolor{NavyBlue}{\textbf{#1}}}
\newcommand{\direc}[1]{\textcolor{ForestGreen}{\textbf{\textit{#1}}}}
\newcommand{\num}[1]{\textcolor{Orange}{\underline{\textbf{#1}}}}

\usepackage{amsmath,amsfonts,bm}

\def\eqref#1{equation~\ref{#1}}
\def\1{\bm{1}}

\def\vv{{\bm{v}}}

\DeclareMathAlphabet{\mathsfit}{\encodingdefault}{\sfdefault}{m}{sl}
\SetMathAlphabet{\mathsfit}{bold}{\encodingdefault}{\sfdefault}{bx}{n}

\def\gX{{\mathcal{X}}}

\title{Diagnosing Vision-and-Language Navigation: What Really Matters}

\author{
Wanrong Zhu\textsuperscript{\P}, 
Yuankai Qi\textsuperscript{\S},
Pradyumna Narayana\textsuperscript{*}, 
Kazoo Sone\textsuperscript{*}, 
Sugato Basu\textsuperscript{*}, \\
\textbf{Eric Xin Wang}\textsuperscript{\ddag}, 
\textbf{Qi Wu}\textsuperscript{\S},
\textbf{Miguel Eckstein}\textsuperscript{\P},
\textbf{William Yang Wang}\textsuperscript{\P} \\
\textsuperscript{\P}UC Santa Barbara, 
\textsuperscript{\S}University of Adelaide,
\textsuperscript{*}Google,
\textsuperscript{\ddag}UC Santa Cruz\\
\texttt{\small \{wanrongzhu,migueleckstein,wangwilliamyang\}@ucsb.edu, qykshr@gmail.com}\\
\texttt{\small \{pradyn,sone,sugato\}@google.com},
\texttt{\small xwang366@ucsc.edu},
\texttt{\small qi.wu01@adelaide.edu.au}
}

\begin{document}
\maketitle

%%%%%%%%% BODY TEXT
%=======================================================================================

%%%%%%%%% ABSTRACT
\begin{abstract}
Vision-and-language navigation (VLN) is a multimodal task where an agent follows natural language instructions and navigates in visual environments. Multiple setups have been proposed, and researchers apply new model architectures or training techniques to boost navigation performance. However, there still exist non-negligible gaps between machines' performance and human benchmarks. Moreover, the agents' inner mechanisms for navigation decisions remain unclear. To the best of our knowledge, how the agents perceive the multimodal input is under-studied and needs investigation. In this work, we conduct a series of diagnostic experiments to unveil agents' focus during navigation. 
Results show that indoor navigation agents refer to both object and direction tokens when making decisions. In contrast, outdoor navigation agents heavily rely on direction tokens and poorly understand the object tokens. 
Transformer-based agents acquire a better cross-modal understanding of objects and display strong numerical reasoning ability than non-Transformer-based agents.
When it comes to vision-and-language alignments, many models claim that they can align object tokens with specific visual targets. We find unbalanced attention on the vision and text input and doubt the reliability of such cross-modal alignments.\footnote{Code and data used in this study are available at \href{https://github.com/VegB/Diagnose_VLN}{https://github.com/VegB/Diagnose\_VLN}.}
\end{abstract}
\section{Introduction}
A key challenge for Artificial Intelligence (AI) research is to move beyond Independent and Identically Distributed (i.i.d.) data analysis: We need to teach AI agents to understand multimodal input data, and jointly learn to reason and perform incremental and dynamic decision-making with the help from humans.
Vision-and-Language Navigation (VLN) has received much attention due to its active perception and multimodal grounding setting, dynamic decision-making nature, rich applications, and accurate evaluation of agents' performances in language-guided visual grounding. As the AI research community gradually shifts its attention from the static empirical analysis of datasets to more challenging settings that require incremental decision-making processes, the interactive task of VLN deserves a more in-depth analysis of why it works and how it works.

Various setups have been proposed to address to the VLN task. Researchers generate visual trajectories and collect human-annotated instructions for indoor~\cite{DBLP:conf/cvpr/AndersonWTB0S0G18,DBLP:conf/acl/JainMKVIB19,DBLP:conf/emnlp/KuAPIB20,chen2021hamt} and outdoor environment~\cite{DBLP:conf/cvpr/ChenSMSA19,DBLP:journals/corr/abs-2001-03671,DBLP:conf/nips/MirowskiGMHATSK18}. There are also interactive VLN settings based on dialogues~\cite{DBLP:conf/cvpr/NguyenDBD19,DBLP:conf/emnlp/NguyenD19,DBLP:conf/cvpr/0004ZZLJCL20}, and task that navigates agents to localize a remote object~\cite{DBLP:conf/cvpr/QiW0WWSH20}. 
However, few studies ask the \textit{Why} and \textit{How} questions: Why do these agents work (or do not work)? How do agents make decisions in different setups?

\begin{table}[t!]
% \vspace{-1ex}
\begin{adjustbox}{width=\linewidth,center}
\begin{tabular}{l  r r r}
\cmidrule[\heavyrulewidth]{1-4}
 &   \textbf{R2R}   & \textbf{RxR} & \textbf{Touchdown}\\
\cmidrule[\heavyrulewidth]{1-4}    
Human Performance       & 86    & 94    & 92 \\
SoTA Model Performance  & 78    & 53    & 17 \\     
\cmidrule[\heavyrulewidth]{1-4}
\end{tabular}
\end{adjustbox}
\caption{There exists salient gaps between machines' vision-and-language navigation (VLN) performance and human benchmarks. Navigation success rates are reported on the R2R~\citep{DBLP:conf/cvpr/AndersonWTB0S0G18} and the RxR dataset~\cite{rxr} for indoor VLN and the Touchdown dataset~\citep{DBLP:conf/cvpr/ChenSMSA19} for outdoor VLN.\protect\footnotemark}
% \vspace{-5pt}
\label{tab:human_sota_comparison}
\end{table}

\addtocounter{footnote}{0}
\footnotetext{We record the published state-of-the-art performance on R2R, RxR and Touchdown leaderboards on Dec.15th, 2021. }

Through the years, agents with different model architectures and training mechanisms have been proposed for indoor VLN~\cite{DBLP:conf/cvpr/AndersonWTB0S0G18,DBLP:conf/nips/FriedHCRAMBSKD18,DBLP:conf/cvpr/HaoLLCG20,DBLP:conf/nips/HongOQ0G20,DBLP:journals/corr/abs-2011-13922,DBLP:conf/iccv/HuangJMKMBI19,DBLP:conf/cvpr/KeLBHGLGCS19,DBLP:conf/emnlp/LiLXBCGSC19,DBLP:conf/iclr/MaLWAKSX19,DBLP:conf/eccv/QiPZHW20,DBLP:conf/naacl/TanYB19,DBLP:conf/eccv/WangWS20,DBLP:conf/cvpr/WangHcGSWWZ19,DBLP:conf/eccv/WangXWW18,DBLP:conf/eccv/WangJIWKR20,DBLP:conf/cvpr/Zhu0CL20} and outdoor VLN~\cite{DBLP:conf/cvpr/ChenSMSA19,DBLP:conf/cvpr/MaWAXK19,DBLP:conf/nips/MirowskiGMHATSK18,DBLP:journals/corr/abs-2003-00857,DBLP:conf/emnlp/Xiang0W20,DBLP:journals/corr/abs-2007-00229}. Back-translation eases the urgent problem of data scarcity~\cite{DBLP:conf/nips/FriedHCRAMBSKD18}. Imitation learning and reinforcement learning enhance agents' generalization ability~\cite{DBLP:conf/cvpr/WangHcGSWWZ19,DBLP:conf/eccv/WangXWW18}. With the rise of BERT-based models, researchers also apply Transformer and pre-training to further improve navigation performance~\cite{DBLP:conf/cvpr/HaoLLCG20,DBLP:journals/corr/abs-2011-13922,DBLP:journals/corr/abs-2007-00229}. 
While applying new techniques to the navigation agents might boost their performance, we still know little about how agents make each turning decision. 
Treatment of the agents' processing of instructions and perception of the visual environment as a black box might hinder the design of a generic model that fully understands visual and textual input regardless of VLN setups.
Table~\ref{tab:human_sota_comparison} shows non-negligible performance gaps between neural agents and humans on both indoor and outdoor VLN tasks.

Therefore, we focus on analyzing how the navigation agents understand the multimodal input data in this work. We conduct our investigation from the perspectives of natural language instruction, visual environment, and the interpretation of vision-language alignment. 
We create counterfactual interventions to alter the instructions and the visual environment in the validation dataset, focusing on variables related to objects, directions and numerics. 
More specifically, we modify the instruction by removing or replacing the object/direction/numeric tokens, and we adjust the environment by masking out visual instances or horizontally flipping the viewpoint images. 
Subsequently, we examine the interventions' treatment effects on agents' evaluation performance while keeping other variables unchanged.
We set up experiments on the R2R\mbox{~\citep{DBLP:conf/cvpr/AndersonWTB0S0G18}} and the RxR dataset\mbox{~\cite{rxr}} for indoor VLN and the Touchdown dataset\mbox{~\citep{DBLP:conf/cvpr/ChenSMSA19}} for outdoor VLN. We examine nine VLN agents on the three datasets with quantitative ablation diagnostics on the text and visual inputs.

In summary, our key findings include: 

\begin{enumerate}[noitemsep, topsep=0pt]
    \item Indoor navigation agents refer to both objects and directions in the instruction when making decisions. In contrast, outdoor navigation agents heavily rely on directions and poorly understand visual objects. (Section~\ref{sec:instr})
    \item Instead of merely staring at surrounding objects, indoor navigation agents are able to set their sights on objects further from the current viewpoint.
    (Section~\ref{sec:visual_env})
    \item Transformer-based agents display stronger numerical reasoning ability (Section~\ref{sec:instr}), and acquire better cross-modal understanding of objects, compared to non-Transformer-based agents. 
    (Section~\ref{sec:vision_language_alignment})
    \item Indoor agents can align object tokens to certain targets in the visual environment to a certain extent, but display in-balanced attention on text and visual input.
    (Section~\ref{sec:vision_language_alignment})
\end{enumerate}

We hope these findings reveal opportunities and obstacles of current VLN models and lead to new research directions.

\section{Related Work}
\paragraph{Instruction Following} is a long-standing topic in AI studies that ask an agent to  follow natural language instructions and accomplish target tasks, which can be dated back to the SHRLDU~\citep{Winograd1971ProceduresAA}. Efforts made to tackle this classic problem spans from defining templates~\citep{Klingspor97human-robot-communicationand,Antoniol2011RobustSU}, designing hard-encoded concepts to ground visual attributes and spatial relations~\citep{Steels1997GroundingAL,Roy2002LearningVG,Guadarrama2013GroundingSR,Kollar2013TowardIG,Matuszek2014LearningFU}, to constructing varies datasets and learning environments~\citep{Anderson1991TheHM,Bisk2016NaturalLC,Misra2018MappingIT}.
% \TODO{model design}
Many methods have been proposed to map the instructions into sequence of actions, such as reinforcement learning~\citep{branavan-etal-2009-reinforcement,branavan-etal-2010-reading,vogel-jurafsky-2010-learning,Misra2017MappingIA}, semantic parsing~\citep{Chen2011LearningTI, Artzi2013WeaklySL}, alignment-based model~\citep{andreas-klein-2015-alignment}, and neural networks~\citep{bisk-etal-2016-natural,Mei2016ListenAA,Tan2018SourceTargetIM}.

\paragraph{Vision-and-Language Navigation} is a task where an agent comprehends the natural language instructions and reasons through the visual environment. 
Many studies aim at improving VLN agents' performance in one way or another. 
To enrich training data, a line of work~\cite{DBLP:conf/nips/FriedHCRAMBSKD18,DBLP:journals/corr/abs-2007-00229} use back-translation to generate augmented instructions.
To enforce cross-modal grounding, RPA and RCM~\cite{DBLP:conf/eccv/WangXWW18,DBLP:conf/cvpr/WangHcGSWWZ19} use reinforcement learning, SMNA~\cite{DBLP:conf/iclr/MaLWAKSX19} uses a visual-textual co-grounding module to improve cross-modal alignment, RelGraph~\cite{DBLP:conf/nips/HongOQ0G20} uses graphs for task formulation. 
To address the generalizability problem to unseen environment, PRESS~\cite{DBLP:conf/emnlp/LiLXBCGSC19} introduces a stochastic sampling scheme,  EnvDrop~\cite{DBLP:conf/naacl/TanYB19} proposes environment dropout.
To utilize visual information from the environment, AuxRN~\cite{DBLP:conf/cvpr/Zhu0CL20} uses auxiliary tasks to assist semantic information extraction, VLN-HAMT~\cite{chen2021hamt} incorporates panorama history with a hierarchical vision transformer.
FAST~\cite{DBLP:conf/cvpr/KeLBHGLGCS19} makes use of asynchronous search and allows the agent to backtrack if it discerns a mistake after attending to global and local knowledge. 
With the success of BERT-related models in NLP, researchers also start to build Transformer-based navigation agents and add a pre-training process before fine-tuning on the downstream VLN task~\cite{DBLP:conf/cvpr/HaoLLCG20,DBLP:journals/corr/abs-2011-13922,DBLP:journals/corr/abs-2007-00229,chen2021hamt}. 
% The increased model size and additional training phase help improve navigation performance to a certain extent.  

\paragraph{Model Behavior Analysis} 
As multimodal studies gain more and more attention, there are lines of works that focus on explaining models' behaviors to better understand and handle the tasks. Some generate textual explanations by training another model to mimic human explanations~\cite{DBLP:conf/eccv/HendricksARDSD16,DBLP:conf/cvpr/ParkHARSDR18,wu-mooney-2019-faithful}.
Others generate visual explanations with the help of attention mechanism~\cite{DBLP:conf/nips/LuYBP16} or gradient analysis~\cite{DBLP:conf/iccv/SelvarajuCDVPB17}. 
There are also attempts to provide multimodal explanations, e.g., ~\citet{DBLP:conf/emnlp/LiFYML18} breaks up the end-to-end VQA process and examines the intermediate results by extracting attributes from the visual instances. 
Another line of work examines model performance by conducting ablation studies on input data. Recent analyses on language modelling~\citep{OConnor2021WhatCF}, machine translation~\citep{fernandes-etal-2021-measuring}, and instruction following~\citep{dan-etal-2021-generalization} ablate/perturb both training and validation data. A study on multimodal models~\citep{Frank2021VisionandLanguageOV} only applies ablation during evaluation, which is the same as our settings.

%======================================================================================
\section{Background and Research Questions}
We first bring in the task of Vision-and-Language Navigation and introduce the datasets and agents used for comparison. Then we list out the research questions to study in this work.

%----------------------------------------------
\subsection{Vision-and-Language Navigation}

In the vision-and-language navigation task, the navigation agent is asked to find the path to reach the target location following the instructions $\gX$.
The navigation procedure can be viewed as a sequential decision-making process.
At each time step $t$, the visual environment presents an image view $\vv_t$. With reference to the instruction $\gX$ and the visual view $\vv_t$, the agent is expected to choose an action $a_t$ such as \textit{turn left} or \textit{stop}.

%----------------------------------------------
% \subsection{Datasets, Models and Metrics}
% \vspace{-3mm}
\paragraph{Datasets}
We conduct indoor navigation experiments on the Room-to-Room (R2R) dataset~\cite{DBLP:conf/cvpr/AndersonWTB0S0G18} and the Room-across-Room (RxR) dataset~\cite{rxr}, and test outdoor VLN on Touchdown~\cite{DBLP:conf/cvpr/ChenSMSA19}. 
R2R and RxR are built upon real estate layouts and contain separate graphs for each apartment/house. Unlike R2R, which shoots for the shortest path, RxR has longer and more variable paths. R2R only contains English instructions, while RxR also includes instructions in Hindi and Telugu. In this study, we only cover the English subset for RxR, and will refer to it as RxR-en in the following sections.
Navigation in Touchdown occurs in the urban environment, where the viewpoints form a huge connected graph. Compared to indoor environments, Touchdown has more complicated visual environments and a more extensive search space. The evaluation results are reported on the validation unseen sets for R2R and RxR-en and on the test set for Touchdown.

% \vspace{-3mm}
\paragraph{Models}

\begin{table}[t!]
% \vspace{-1ex}
\begin{adjustbox}{width=\linewidth,center}
\begin{tabular}{l  l  c  l}
\cmidrule[\heavyrulewidth]{1-4}
\textbf{Dataset} &   \textbf{Model}   & \textbf{Trans?} & \textbf{Visual Feature}\\
\cmidrule[\heavyrulewidth]{1-4}    
\multirow{4}{*}{R2R} 
    & EnvDrop~\cite{DBLP:conf/naacl/TanYB19} & $\times$ & \multirow{4}{*}{ResNet-152} \\
    & FAST~\cite{DBLP:conf/cvpr/KeLBHGLGCS19} & $\times$ &  \\
    & VLN $\circlearrowright$ BERT~\cite{DBLP:journals/corr/abs-2011-13922}   & \checkmark &     \\
    & PREVALENT~\cite{DBLP:conf/cvpr/HaoLLCG20} & \checkmark &   \\
\cmidrule{1-4}
\multirow{2}{*}{RxR-en} 
    & CLIP-ViL~\citep{Shen2021HowMC}    &  $\times$     & \multirow{2}{*}{CLIP-ViT} \\
    & VLN-HAMT~\citep{chen2021hamt}     & \checkmark &  \\
\cmidrule{1-4}
\multirow{3}{*}{Touchdown}  
    & RCONCAT~\cite{DBLP:conf/cvpr/ChenSMSA19}  &  $\times$  & \multirow{3}{*}{ResNet-18}   \\
    & ARC~\cite{DBLP:conf/emnlp/Xiang0W20} & $\times$  &    \\
    & VLN-Transfomer~\cite{DBLP:journals/corr/abs-2007-00229} & \checkmark &  \\
\cmidrule[\heavyrulewidth]{1-4}
\end{tabular}
\end{adjustbox}
\caption{The VLN datasets and models covered in this study. We record whether the model structure is Transformer-based, and the pre-trained feature extractor used to encode visual environment.}
\label{tab:tasks_datasets}
\end{table}

Table~\ref{tab:tasks_datasets} lists out the models covered in our study.
We use the code and trained checkpoints shared by the authors in the following experiments.

For indoor navigation on R2R, we study a widely adopted base model Envdrop~\cite{DBLP:conf/naacl/TanYB19},
a backtracking framework for self-correction FAST~\cite{DBLP:conf/cvpr/KeLBHGLGCS19},
and two SoTA models VLN $\circlearrowright$ BERT~\cite{DBLP:journals/corr/abs-2011-13922} and PREVALENT~\cite{DBLP:conf/cvpr/HaoLLCG20}. 
The Envdrop introduces environment dropout on top of the Speaker-Follower~\cite{DBLP:conf/nips/FriedHCRAMBSKD18} model, FAST conducts an asynchronous search for backtracking, PREVALENT, and VLN $\circlearrowright$ BERT are Transformer-based agents with pre-trained models. 

For navigation on RxR-en, we examine CLIP-ViL~\citep{Shen2021HowMC} and VLN-HAMT~\citep{chen2021hamt}. CLIP-ViL shares the same model structure with EnvDrop. The only difference is that CLIP-ViL uses CLIP-ViT~\citep{Radford2021LearningTV} to extract visual features, while EnvDrop uses ImageNet ResNet~\citep{Szegedy2017Inceptionv4IA} features. VLN-HAMT incorporates a long-horizon history into decision-making by encoding all the past panoramic observations via a hierarchical vision Transformer.

For outdoor navigation on Touchdown, we consider the common baseline RCONCAT~\cite{DBLP:conf/cvpr/ChenSMSA19}, and two SoTA models ARC~\cite{DBLP:conf/emnlp/Xiang0W20} and VLN-Transfomer~\cite{DBLP:journals/corr/abs-2007-00229}. RCONCAT encodes the trajectory and the instruction in an LSTM-based manner. ARC improves RCONCAT by paying special attention to the stop signals. VLN-Transfomer is a Transformer-based agent that applies pre-training on an external dataset for outdoor navigation in urban areas.

% \vspace{-2mm}
\paragraph{Metrics}
In the following experiments, we evaluate navigation performance with Success Rate (SR) for indoor agents and  Task Completion (TC) rate for outdoor agents. Both SR and TC measure the accuracy of completing the navigation task, reflecting the agents' overall ability to finish navigation correctly. An indoor navigation task is considered complete if the agent's final position locates within 3 meters of the target location. For outdoor navigation, the task is considered complete if the agent stops at the target location or one of its adjacent nodes in the environment graph.

%----------------------------------------------
\subsection{Research Questions}

Current VLN studies have reached their bottleneck as only minor performance improvements have been achieved recently, while a significant gap still exists between machine and human performance. 
This motivates us to find the reasons.

To better understand how VLN agents make decisions during navigation, we conduct a series of experiments on indoor and outdoor VLN tasks, aiming to answer the following questions that might help us locate the deficiencies of current model designs and explore future research directions:
\begin{enumerate}[noitemsep, topsep=0pt]
    \normalem
    \item \emph{What can the agents learn from the instructions? Do they pay more attention to object tokens or directions tokens? Do they have the ability to count? (Section~\ref{sec:instr})}
    \item \emph{What do agents see in the visual environment? Are they staring at the closely surrounded objects or also browsing further layout? Do they focus on individual visual instances or perceive the overall outline? (Section~\ref{sec:visual_env})}
    \item \emph{Can agents match textual tokens to visual entities? How reliable are such connections?  (Section~\ref{sec:vision_language_alignment})}
    \ULforem
\end{enumerate}

\begin{table}[!t]
% \vspace{-7px}
\begin{adjustbox}{width=\linewidth,center}
\begin{tabular}{l c }
\cmidrule[\heavyrulewidth]{1-2}
\textbf{Dataset} & \textbf{Instruction} \\ \cmidrule[\heavyrulewidth]{1-2}

R2R     & \makecell[l]{
Walk through the \obj{door} by the \obj{sink} into the \obj{middle} of the next
\\\obj{room}. Turn \direc{right} and walk down the \obj{hallway} and enter the
\\\num{third} \obj{door} on your \direc{right}.}
\\ \cmidrule{1-2}

RxR-en     &  \makecell[l]{
We're facing towards a small \obj{picture} that's attached to the \obj{wall},
\\turn slightly to the \direc{right}, and enter the \obj{hallway} that's in \direc{front} of
\\you, turn to the \direc{left}, take \num{five} \obj{steps} further... On your \direc{right} there
\\are a few \obj{glass doors} and on the \direc{left} there's a \obj{living room}, walk
\\towards the \obj{living room}, turn slightly to the \direc{left}... On the \direc{right}
\\there are \num{four} \obj{chairs} and a beautiful \obj{coffee table} in the \obj{middle},
\\on the \direc{left} there's a \obj{console table} with a \obj{vase} with \obj{flowers} on
\\\obj{top}, walk past the \obj{console table} towards the \direc{back} of the \obj{chair}
\\that's in direc{front} of you... We're now facing towards a \obj{lamp},
\\and on the \direc{right} there's a marble \obj{console table} with \obj{decorations}
\\on \obj{top}, and that's your \obj{destination}. 
}
\\ \cmidrule{1-2}

Touchdown     & \makecell[l]{
Orient yourself so that you are moving in the same \obj{direction} as
\\\obj{traffic}. Go straight through \num{3} \obj{intersections}. Keep moving \direc{forw-}
\\\direc{ard}, after the \num{3rd} \obj{intersection}, you should see a \obj{signs} for a \obj{store}
\\with a white \obj{background} and red \obj{dots} as well as a red and white
\\\obj{bullseye} \obj{target}. Continue going straight past this \obj{store} and at the
\\next \obj{intersection}, turn \direc{left}. Go through \num{one} \obj{intersection} and \direc{stop}
\\just after the \obj{wall} on your \direc{left} with the purple \obj{zig} \obj{zag} \obj{patterns}.
}
\\ \cmidrule[\heavyrulewidth]{1-2}

\end{tabular}
\end{adjustbox}
\caption{Instructions from R2R, RxR-en and Touchdown with \obj{object-tokens}, \direc{direction-tokens} and \num{numeric-tokens} highlighed.
}
% \vspace{-15px}
\label{tab:instr_examples}
\end{table}

\begin{table}[!t]
\begin{adjustbox}{width=\linewidth,center}
\begin{tabular}{lcrrccc}
\cmidrule[\heavyrulewidth]{1-7}
\textbf{Dataset} & \textbf{\#Data}  & \textbf{$\overline{\text{L}_\text{path}}$} & \textbf{$\overline{\text{L}_\text{instr}}$} & \textbf{\#Object} & \textbf{$\overline{p(\text{tok}_\text{obj})}$} & \textbf{$\overline{p(\text{tok}_\text{direc})}$} \\ \cmidrule[\heavyrulewidth]{1-7}

R2R         & 2.3k    &   6.0 &   29.3    &   0.6k    &   19.8\%  &   7.3\%       \\ 
RxR-en         & 4.6k    &   8.5 &   111.3    &   1.4k    &   16.1\%  &   6.5\%       \\ 
Touchdown   &   1.4k    &   34.4    &   92.5    &   1.0k    &   16.8\%  &   6.8\%      \\ 
\cmidrule[\heavyrulewidth]{1-7}
\end{tabular}
\end{adjustbox}
\caption{Statistics of R2R, RxR-en and Touchdown datasets.
\#Data is the dataset size used for evaluation in this study.
$\overline{L_\text{path}}$ is the average path length, which is the number of viewpoints covered in the trajectory.
$\overline{L_\text{instr}}$ is the average instruction length.
\#Object denotes the number of unique objects mentioned in the instructions. 
$\overline{p(\text{tok}_\text{obj})}$ and $\overline{p(\text{tok}_\text{direc})}$ indicates the percentage of object/direction tokens per instruction.
}
% \vspace{-2mm}
\label{tab:dataset_comparison}
\end{table}

\section{Analysis on Instruction Understanding}
\label{sec:instr}

\begin{table*}[!t]
\begin{adjustbox}{width=\linewidth,center}
\begin{tabular}{l l l }
\cmidrule[\heavyrulewidth]{1-3}
\textbf{*} & \textbf{Setting} & \multicolumn{1}{c}{\textbf{Instruction}} \\ \cmidrule[\heavyrulewidth]{1-3}
1   &   Vanilla     & \makecell[l]{Go \direc{left} down the \obj{hallway} toward the \obj{exit sign}. Go into the \obj{door} on the \direc{left} and \direc{stop} by the \obj{table}. }\\ \cmidrule[\heavyrulewidth]{1-3}

2   &   Mask Object Tokens     & \makecell[l]{Go left down the \obj{[MASK]} toward the \obj{[MASK]} \obj{[MASK]}. To into the \obj{[MASK]} on the left and stop by the  \obj{[MASK]}.}\\ 
% \cmidrule{1-2}
3   &   Replace Object Tokens     & \makecell[l]{Go left down the \obj{portrait} toward the \obj{sofa} \obj{fountains}.  Go into the \obj{football} on the left and stop by the \obj{boats}.}\\ 
% \cmidrule{1-2}
4   &   Controlled Trial    & \makecell[l]{Go  \obj{[MASK]} down the hallway \obj{[MASK]} the exit sign. To into the door on \obj{[MASK]} left and  \obj{[MASK]} by  \obj{[MASK]} table.}\\
\cmidrule[\heavyrulewidth]{1-3}

5   &   Mask Direction Tokens     & \makecell[l]{Go \direc{[MASK]} down the hallway toward the exit sign. Go into the door on the \direc{[MASK]} and \direc{[MASK]} by the table.}\\ 
% \cmidrule{1-2}
6   &   Replace Direction Tokens     & \makecell[l]{Go \direc{right} down the hallway toward the exit sign.  Go into the door on the \direc{right} and \direc{forward} by the table.}\\ 
% \cmidrule{1-2}
7   &   Controlled Trial    & \makecell[l]{Go left down the \direc{[MASK]} \direc{[MASK]} the exit sign.  Go into the door on the left and \direc{[MASK]} by the table.}\\ \cmidrule[\heavyrulewidth]{1-3}
\end{tabular}
\end{adjustbox}
\caption{ Example of instruction modification. 
In the original instruction, there are five \obj{object-related tokens}, and three \direc{direction-related tokens}. In the object token ablations, we mask out the object tokens, or replace them with randomly sampled object tokens. The controlled trial randomly masked out five tokens from the instruction for a fair comparison.
Likewise the direction tokens.
}
% \vspace{-4mm}
\label{tab:masking_examples}
\end{table*}

\begin{table*}[htbp]
% \vspace{-1ex}
\begin{adjustbox}{width=\linewidth,center}
\begin{tabular}{l|l | l | r r r r |  r r | r r  r}
\cmidrule[\heavyrulewidth]{1-12}
\multirow{2}{*}{\textbf{*}}    & \multirow{2}{*}{\textbf{Ablation}}    & \multirow{2}{*}{\textbf{Setting}}  & \multicolumn{4}{c}{\textbf{SR $\uparrow$ on R2R} } & \multicolumn{2}{|c}{\textbf{SR $\uparrow$ on RxR-en}  } & \multicolumn{3}{|c}{\textbf{TC $\uparrow$ on Touchdown}}\\ \cmidrule{4-12}

    &    &   & EnvDrop   & FAST  & VLN$\circlearrowright$BERT   & PREVALENT & CLIPViL & HAMT & RCONCAT   & ARC   & VLNTrans \\ \cmidrule[\heavyrulewidth]{1-12}
	
1   &	~~~~~~--	&	Vanilla	&	49.77~	&	63.90~	&	53.30~	&	57.13~	&	40.21~	&	52.52~	&	11.78~	&	15.19~	&	16.11~	\\ 
\cmidrule[\heavyrulewidth]{1-12}
2   &   \multirow{2}{*}{Object}	&	Mask	&	-36\%	&	-38\%	&	-21\%	&	-20\%	&	-48\%	&	-32\%	&	-34\%	&	-36\%	&	-6\%	\\
3   &    &   Controlled Trial	&	-31\%	&	-26\%	&	-8\%	&	-8\%	&	-35\%	&	-23\%	&	-44\%	&	-55\%	&	-15\%	\\
\cmidrule[\heavyrulewidth]{1-12}
4   &   \multirow{2}{*}{Direction}	&	Mask	&	-23\%	&	-23\%	&	-15\%	&	-11\%	&	-39\%	&	-28\%	&	-73\%	&	-90\%	&	-45\%	\\
5   &    &   Controlled Trial	&	-11\%	&	-12\%	&	-4\%	&	-3\%	&	-14\%	&	-9\%	&	-22\%	&	-23\%	&	-8\%	\\ 
\cmidrule[\heavyrulewidth]{1-12}    
\end{tabular}
\end{adjustbox}
\caption{The navigation performance for indoor and outdoor agents on object-token and direction-token ablations. 
We record the validation score in the ``vanilla'' setting, and report the relative performance change for each ablation setting.
For object-token ablations, the ``mask'' setting masks out all the object-tokens, while the controlled trial masks out the same amount of random tokens. The same applies to direction-token ablations.
}
\label{tab:instr_obj_direc_tokens}
\end{table*}

This section examines whether and to what extent the agent understands VLN instructions. 
We focus on how the agent perceives object-related tokens, direction-related tokens, and numeric tokens, and their effects on final navigation performance. 
Table~\ref{tab:instr_examples} shows exemplar instructions of the three datasets covered in our study. 
As shown in Table~\ref{tab:dataset_comparison}, Touchdown's trajectory length is significantly longer than the other two indoor datasets. RxR-en and Touchdown have longer instructions than R2R. The ratios of object and direction tokens in all three datasets are comparable, involving about two times more object tokens than direction tokens.

%----------------------------------------------
\subsection{The Effect of Object-related Tokens}
\label{sec:instr_object}

We first create counterfactual interventions on instructions by masking out the object tokens. We use Stanza~\cite{DBLP:conf/acl/QiZZBM20} part-of-speech (POS) tagger to locate object-related tokens. A token will be regarded as an object token if its POS tag is \textit{NOUN} or \textit{PROPN}. During masking, we replace the object token with a specified mask token \textit{[MASK]}. Then we examine the average treatment effects of the intervention on agents' performance, while keeping other variables unchanged. 

Noticeably, when we mask out the object tokens, the number of visible tokens in the provided instruction also decreases, which is a coherent factor with masking object tokens and might interfere with our analysis. To eliminate the effect of reducing visible tokens, we add a controlled trial in which we randomly mask out the same amount of tokens. Table~\ref{tab:masking_examples} gives an example of masking object tokens (\#2) and its corresponding controlled trial (\#4).

We follow each agent's original experiment setting for all the experiments in this study and train it on the original train set. Then we apply masking to object tokens in the validation set, and report agents' relative performance changes under each setting. We conduct five repetitive experiments and report the average scores for settings that involve random masking or replacing.

Table~\ref{tab:instr_obj_direc_tokens} presents how the agents' navigation performance change when object tokens are masked out (\#2 \& \#3). Intuitively, not knowing what objects are mentioned in the instruction lowers all models' performance. 
Comparing the masking ablations with the controlled trial for indoor VLN, we notice that masking out the object tokens result in a more drastic decrease in success rate than masking out random tokens. This holds for all indoor agents, which verifies that {\uline{indoor agents depend on object tokens more than other tokens}}. 
However, when we compare results on the Touchdown for outdoor VLN, we notice in surprise that masking out the object tokens has a weaker impact on task completion rate than masking out random tokens. This suggests that current {\uline{outdoor navigation agents do not fully take object tokens into consideration}} when making decisions. This may be caused by the weak visual recognition module in current outdoor agents. As addressed in Table\mbox{~\ref{tab:tasks_datasets}}, all three outdoor agents rely on visual features extracted by ResNet-18, which may not be powerful enough to fully incorporate the complicated urban environments.

%----------------------------------------------
\subsection{The Effect of Direction-related Tokens}
\label{sec:instr_direction}

We regard the following tokens as direction-related tokens: \textit{left, right, back, front, forward, stop}. Similar to how we ablate the object tokens, we mask out direction tokens from the instruction and examine the impact on agents' navigation performance. Table~\ref{tab:masking_examples} provides examples of direction tokens masking (\#5), and its controlled trial (\#7) where the same amount of random tokens are masked out.
Table~\ref{tab:instr_obj_direc_tokens} shows agents' performance under various direction tokens ablation settings (\#4 \& \#5). 

For indoor agents, masking out the direction tokens cause a sharper drop in success rate compared to masking out random tokens, which means the indoor navigation agents do consider the direction tokens during navigation. 
We also notice that agents are more sensitive to the loss of direction guidance on RxR-en than on R2R. Such difference may be caused by the way these two datasets are designed. R2R's ground-truth trajectories are the shortest path from start to goal. Previous studies have noted that R2R has the danger of exposing structural bias and leaking hidden shortcuts\mbox{~\citep{Thomason2019ShiftingTB}}, and that such design encourages goal-seeking over path adherence\mbox{~\citep{Jain2019StayOT}}. RxR is crafted to include longer and more variable paths to avoid such biases. Naturally, agents on RxR-en would pay more attention to direction tokens since they may approach their goal indirectly.

For outdoor navigation agents, masking out direction tokens leads to a drastic decline in task completion rate, compared to random masking. This indicates that current {\uline{outdoor navigation agents heavily rely on the direction tokens}} when making decisions. Given the complicated visual environments and instructions in the outdoor navigation task, current agents fail to fully use the instructions, especially ignoring the rich object-related information. 
The ARC model shows the most salient performance decline of 90\% to the instructions ablated by direction token masking. Aside from the classifier that predicts the next direction to take, ARC also uses a stop indicator to decide whether to stop at each step or not. Its unique mechanism for detecting stop signals might explain why it is more sensitive to the existence of direction tokens.

\begin{table}[!t]
% \vspace{-1ex}
\begin{adjustbox}{width=0.7\linewidth,center}
\begin{tabular}{l | r r r}
\cmidrule[\heavyrulewidth]{1-4}
    & \textbf{\#Data}    &  \textbf{$\overline{\text{L}_\text{instr}}$}  &   \textbf{$\overline{p(\text{tok}_\text{num})}$}  \\ 
\cmidrule[\heavyrulewidth]{1-4} 
RxR-en     &   2.0k    & 135.0 &   1.4\% \\
Touchdown   & 1.0k  & 100.1  &   2.0\% \\
\cmidrule[\heavyrulewidth]{1-4}    
\end{tabular}
\end{adjustbox}
\caption{Statistics of RxR-en and Touchdown data samples with numeric tokens examined for evaluation.  $\overline{\text{L}_\text{instr}}$ is the average instruction length, and  $\overline{p(\text{tok}_\text{num})}$ denotes the percentage of numeric tokens per instruction.
}
\label{tab:num_instr_statistics}
\end{table}

%----------------------------------------------
\subsection{The Effect of Numeric Tokens}
\label{sec:instr_count}

We conduct ablation studies on agents' understanding of numeric tokens on RxR-en for indoor agents and Touchdown for outdoor agents. We select a subset of examples whose instructions contain numeric tokens,\footnote{We consider instructions that contain cardinal numbers from 1 to 20, and ordinal numbers from 1st to 20th.} and construct ablated instructions on top. Table~\ref{tab:num_instr_statistics} provides the statistics of the instructions for numeric ablations. Table~\ref{tab:instr_count_results} lists out the results. 
The VLN-HAMT on RxR-en and VLN-Transformer on Touchdown have comparable performance when masking numeric tokens over random tokens, and have worse performance when replacing numeric tokens. This suggests that these two Transformer-based agents have the ability to conduct numerical reasoning to some extent. 
In contrast, other non-Transformer-based agents have less salient performance drops when replacing numeric tokens. For RCONCAT and ARC, replacing numeric tokens even leads to higher task completion rates. This implies the insufficient counting ability of the non-Transformer-based agents.

\begin{table}[!t]
% \vspace{-2ex}
\begin{adjustbox}{width=\linewidth,center}
\begin{tabular}{l | r r | r r r}
\cmidrule[\heavyrulewidth]{1-6}
 \multirow{2}{*}{\textbf{Setting}}  & \multicolumn{2}{c}{\textbf{SR $\uparrow$ on RxR-en} } & \multicolumn{3}{|c}{\textbf{TC $\uparrow$ on Touchdown} } \\ \cmidrule{2-6}
     &  CLIPViL & HAMT & RCONCAT & ARC & VLNTrans\\
\cmidrule[\heavyrulewidth]{1-6}
Vanilla	&	36.05~	&	47.38~	&	11.76~	&	14.24~	&	16.31~	\\
\cmidrule{1-6}
Mask Number	&	-4\%	&	-5\%	&	3\%	&	-7\%	&	-3\%	\\
Replace Number	&	-3\%	&	-6\%	&	1\%	&	2\%	&	-11\%	\\
Controlled Trial	&	-6\%	&	-3\%	&	-5\%	&	-4\%	&	-6\%	\\
\cmidrule{1-6}
\cmidrule[\heavyrulewidth]{1-6}    
\end{tabular}
\end{adjustbox}
\caption{Navigation performance on different numeric-token ablations settings.}
\label{tab:instr_count_results}
\end{table}

\section{Analysis on Visual Environment}
\label{sec:visual_env}

This section investigates what the agent perceives in the visual environment. We set an eye on inspecting the agent's understanding of the surrounding objects and direction-related information.

% ----------------------------------------------
\subsection{Effect of Objects in the Environment}
\label{sec:env_object}

Built upon the Matterport dataset~\cite{DBLP:conf/3dim/ChangDFHNSSZZ17}, R2R and RxR obtain detailed object instance annotations and serve as an excellent source for our visual object studies. Touchdown is based on Google Street View and does not acquire object-related annotations. Thus, we conduct experiments on the indoor VLN environment.

\begin{figure}[!t]
\begin{subfigure}[t]{0.49\linewidth}
    \includegraphics[width=\linewidth]{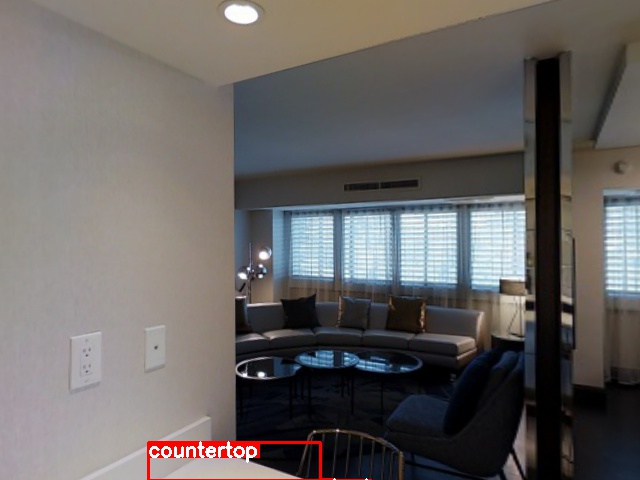}
\caption{Foreground Objects}
\label{fig:eg_3m}
\end{subfigure}\hfill
\begin{subfigure}[t]{0.49\linewidth}
    \includegraphics[width=\linewidth]{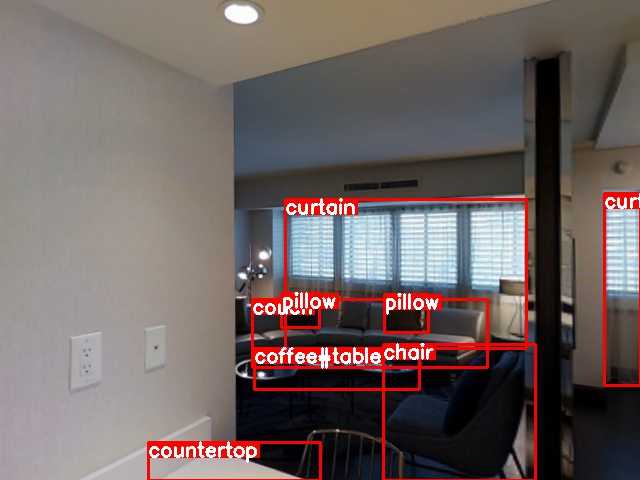}
\caption{All Visible Objects}
\label{fig:eg_ex_wfc}
\end{subfigure}\hfill

\caption{
Accessible objects within different ranges.
}
% \vspace{-7mm}
\label{fig:mask_env_eg}
\end{figure}

We designed several ablation settings for visual objects. The ``mask all visible'' setting applies masking to all the visible visual objects in the environment (except for wall/ceiling/floor). The ``mask foreground'' setting ablates the visual objects within 3 meters of the camera viewpoint, which we refer to as the foreground area. The region beyond 3 meters from the camera viewpoint is regarded as the background area. 
Figure\mbox{~\ref{fig:mask_env_eg}} shows an example for comparison. 
We choose 3 meters as the boundary because the bounding box annotations for objects within 3 meters are provided in REVERIE\mbox{~\cite{DBLP:conf/cvpr/QiW0WWSH20}}. We denote the number of visual objects within 3 meters as $k$, and add a controlled trial that masks out $k$ random visual objects from all the visible objects at the current viewpoint, regardless of their depth. 

Table\mbox{~\ref{tab:mask_env_statistics}} compares the number of visual objects under various ablation settings. 
We mask out the objects in each view by filling the corresponding bounding boxes with the mean color of the surrounding. 
Then we follow original experiment settings and use ResNet-152~\cite{DBLP:conf/cvpr/HeZRS16} CNN to extract image features for R2R agents, and use CLIP-ViT-B/32~\citep{Radford2021LearningTV} to extract visual features for RxR-en agents.

\begin{table}[!t]
% \vspace{-1ex}
\small
\begin{adjustbox}{width=\linewidth,center}
\begin{tabular}{lcr}
\cmidrule[\heavyrulewidth]{1-3}
\textbf{Setting} & \textbf{R2R} & \textbf{RxR-en}  \\ \cmidrule[\heavyrulewidth]{1-3}

All Visible Objects (except wall/floor/ceiling)        & 35.1  &    35.5    \\ 
% \cmidrule{1-2}
Foreground Objects         &  3.2   & 3.9   \\ 
% \cmidrule{1-2}
Objects Mentioned in Instruction         &  5.3     & 11.8  \\   \cmidrule[\heavyrulewidth]{1-3}
\end{tabular}
\end{adjustbox}
\caption{The average number of visual objects in the panorama at each viewpoint under different settings.}
% \vspace{-15px}
\label{tab:mask_env_statistics}
\end{table}

Results for visual object ablations are shown in Table\mbox{~\ref{tab:env_obj_direct_results}}.
We examine the influence of masking out different quantities of visual objects by comparing the ``mask all visible'' setting with the controlled trial (\#2 vs. \#4). It comes naturally that masking out all the visible objects has a more salient impact on the success rate for all the listed indoor agents.
We study the influence of masking visual objects at different depths by comparing the ``mask foreground'' setting with the controlled trial (\#3 vs. \#4). 
Noted here that the number of foreground objects is limited. Thus only a few objects are being masked out in both settings. Still, all listed indoor agents display worse performance on the controlled trial. Such results state that masking out further visual instances in the background, even only a tiny amount, will hurt navigation performance. This indicates that the tested agents consider all the objects in the visual environment during navigation, instead of merely staring at the closely surrounding objects. 

Notice that the agent designs, the dataset domains, and the visual feature extractors are three coherent factors that may result in performance differences. We further justify this by adding another set of ablation studies, where we apply ImageNet ResNet-152 and CLIP-ViT to extract visual features for R2R and RxR-en, and evaluate with the same agent model EnvDrop. Results are shown in Table~\ref{tab:envdrop_ablation}. The trend of different masking settings aligns with our previous findings in Table~\ref{tab:env_obj_direct_results}, and verifies that the background information is also crucial in the visual features.

\begin{table}[!t]
% \vspace{-2ex}
\begin{adjustbox}{width=\linewidth,center}
\begin{tabular}{l| l | l | r r r r | r r}
\cmidrule[\heavyrulewidth]{1-9}
 \multirow{2}{*}{\textbf{*}} & \multirow{2}{*}{\textbf{Ablation}}  & \multirow{2}{*}{\textbf{Setting}}  & \multicolumn{4}{c}{\textbf{SR $\uparrow$ on R2R} } & \multicolumn{2}{|c}{\textbf{SR $\uparrow$ on RxR-en} } \\ \cmidrule{4-9}
    &    &     & EnvDrop   & FAST  & Recur   & PVLT & CLIPViL & HAMT \\
\cmidrule[\heavyrulewidth]{1-9}
1   &   -   & Vanilla	&	49.77~	&	63.90~	&	53.30~	&	57.13~	&	40.21~	&	52.52~	\\
\cmidrule{1-9}
2   &    \multirow{3}{*}{Object}  & MAV	&	-34\%	&	-67\%	&	-37\%	&	-47\%	&	-30\%	&	-43\%	\\
3   &    &   MFG	&	-3\%	&	-6\%	&	-1\%	&	-8\%	&	-2\%	&	-2\%	\\
4   &    &   CT	&	-5\%	&	-10\%	&	-6\%	&	-9\%	&	-3\%	&	-5\%	\\
\cmidrule{1-9}
5   &   \multirow{1}{*}{Direction}	&	Flip	&	-41\%	&	-30\%	&	-48\%	&	-59\%	&	-36\%	&	-47\%	\\
\cmidrule[\heavyrulewidth]{1-9}    
\end{tabular}
\end{adjustbox}
\caption{Indoor navigation performance on various ablation settings on the visual environment.
We compare three masking settings on the visual objects: mask all visible objects (MAV), mask foreground objects (MFG), and the controlled trial (CT).
We horizontally flip the viewpoint to ablate direction-related visual information.
Recur: VLN $\circlearrowright$ BERT. PVLT: PREVALENT.
}
\label{tab:env_obj_direct_results}
\end{table}

\begin{table}[!t]
\vspace{-1ex}
\begin{adjustbox}{width=0.9\linewidth,center}
\begin{tabular}{l l r r r r}
\cmidrule[\heavyrulewidth]{1-6}
\textbf{Dataset}  & \textbf{Visual Feature} & \textbf{Vanilla} & \textbf{MAV}    & \textbf{MFG} & \textbf{CT} \\ \cmidrule[\heavyrulewidth]{1-6}
\multirow{2}{*}{R2R} & ResNet-152    &	49.77	&	-34\%	&	-3\%	&	-5\%	\\
     & CLIP-ViT    &	56.36	&	-34\%	&	-2\%	&	-5\%	\\
\cmidrule{1-6}    
\multirow{2}{*}{RxR-en}  & ResNet-152    &	35.27	&	-42\%	&	-1\%	&	-3\%	\\
     & CLIP-ViT      &	40.21	&	-30\%	&	-2\%	&	-3\%	\\
\cmidrule[\heavyrulewidth]{1-6}    
\end{tabular}
\end{adjustbox}
\caption{EnvDrop's navigation performance on R2R and RxR-en with different visual object ablation settings when using different visual features. MAV: mask all visible objects. MFG: mask foreground objects. CT: controlled trial.}
\label{tab:envdrop_ablation}
\end{table}

\subsection{Effect of Directions in the Environment}
\label{sec:env_direction}

In this ablation setting, we randomly flip some of the viewpoints horizontally.
The objects' relative positions at the flipped viewpoints will be reversed. Presumably, suppose the agent can follow the instruction and find the corresponding direction to approach. In that case, the flipped viewpoints will misguide the agent in the opposite direction and lower the navigation success rate.
As shown in Table~\ref{tab:env_obj_direct_results}, flipping the viewpoints leads to drastic declines in the success rate for all listed indoor agents. This verifies our previous finding that indoor agents can understand directions in the instruction.
We notice that FAST is the only listed model that is less affected by the direction flipping ablation than by the object masking ablation (\#2 vs. \#5). This suggests that FAST's asynchronous backtracking search is able to adjust and recover from errors that occur when choosing directions to some extent.
\section{Analysis on Vision-Language Alignment}
\label{sec:vision_language_alignment}

This section examines the agents' ability to learn vision-language alignment when executing the navigation. We focus on whether the agents can understand the objects mentioned in the instruction and align them to the correct visual instance in the environment, which is crucial to completing this multimodal task. To verify the existence of vision-language alignment, we add perturbations to the visual and textual input, and check how they affect agents' performance.

%----------------------------------------------
\subsection{Instruction Side Perturbation}
\label{sec:align_instr}

\begin{figure}[!t]
% \vspace{-5pt}
    \centering
    \includegraphics[width=0.95\linewidth]{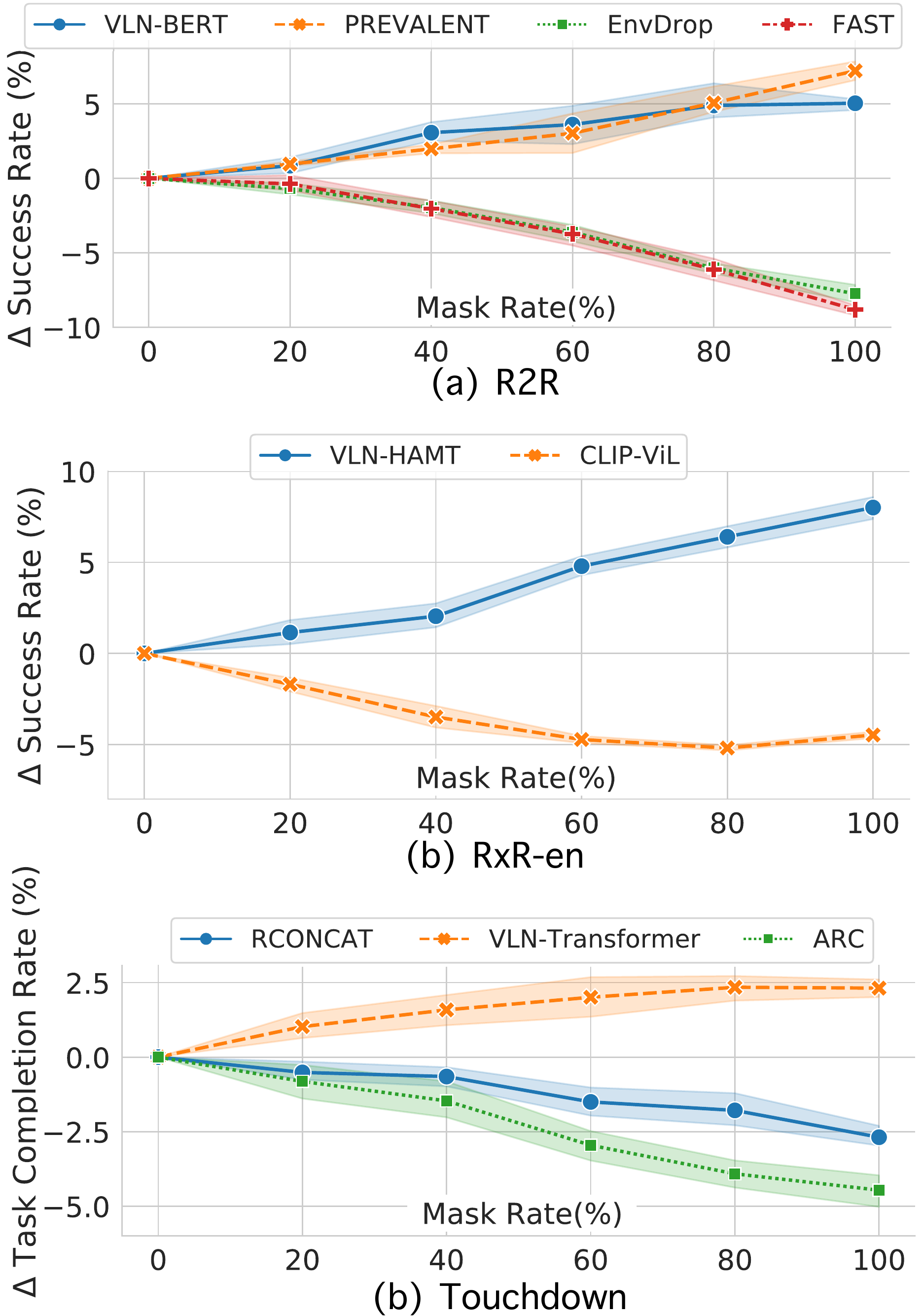}
    \caption{Performance gap between masking and replacing object tokens from instructions. If $\Delta > 0$, then replacing object tokens leads to worse navigation performance, which suggests that the agent has a better understanding of the object tokens.}
    \label{fig:instr_mask_replace_delta}
\end{figure}

We add noise to the textual input by randomly replacing object tokens with random object tokens in the instruction. Table~\ref{tab:masking_examples} shows an example (\#3). This experiment aims to verify whether the agent can line the object tokens up to certain visual targets.
 The assumption is that if the agent can correctly align objects mentioned in the instruction to some targets in the visual environment, then replacing the object token will confuse and misguide the agent. 
Examining Figure~\ref{fig:instr_mask_replace_delta}, we notice that for all three datasets, the Transformer-based models have worse performance when replacing the object tokens, compared to simple masking. This indicates that {\uline{Transformer-based models have a better cross-modal understanding of objects, and can align object tokens to the visual targets}}.
Such superior performance may result from the fact that the Transformer-based models are often pre-trained on multimodal resources, thus displaying a slightly more vital ability to form alignment.

%----------------------------------------------
\subsection{Environment Side Perturbation}
\label{sec:align_env}

\begin{table}[!t]
% \vspace{-1ex}
\begin{adjustbox}{width=\linewidth,center}
\begin{tabular}{l | r r r r | r r}
\cmidrule[\heavyrulewidth]{1-7}
\multirow{2}{*}{\textbf{Setting}}  & \multicolumn{4}{c}{\textbf{SR $\uparrow$ on R2R} } & \multicolumn{2}{|c}{\textbf{SR $\uparrow$ on RxR-en} } \\ \cmidrule{2-7}
     & EnvDrop   & FAST  & Recur   & PVLT & CLIPViL & HAMT \\
\cmidrule[\heavyrulewidth]{1-7}
Vanilla	&	49.77~	&	63.90~	&	53.30~	&	57.13~	&	40.21~	&	52.52~	\\
\cmidrule{1-7}
Dynamic Mask	&	-10\%	&	-24\%	&	-5\%	&	-13\%	&	-19\%	&	-23\%	\\
Controlled Trial	&	-8\%	&	-19\%	&	-3\%	&	-11\%	&	-16\%	&	-20\%	\\
\cmidrule{1-7}
Mask Tokens	&	-36\%	&	-38\%	&	-21\%	&	-20\%	&	-48\%	&	-32\%	\\
\cmidrule[\heavyrulewidth]{1-7}    
\end{tabular}
\end{adjustbox}
\caption{Indoor navigation performance when dynamically masking out the visual objects mentioned in the instructions. Recur: VLN $\circlearrowright$ BERT. PVLT: PREVALENT.
}
\label{tab:mask_env_dynamic}
\end{table}

We add noise to the visual input by conducting the following ablations. In the ``dynamic mask'' setting, we dynamically mask out the visual object regions mentioned in the instruction. We randomly mask out the same amount of visual objects at each viewpoint in its controlled trial. We also compare with the ``mask tokens'' setting, where we mask out all the object tokens in the instruction, while leaving the visual environment untouched. 
This experiment aims to determine if the agent aligns the textual object tokens to the correct visual target. The assumption is that if the agent builds proper vision-language alignment and we mask out visual objects mentioned in the instruction, then the agent may get confused since it can not find the counterpart in the visual environment.

Results are shown in Table\mbox{~\ref{tab:mask_env_dynamic}}. The success rate witnesses a decline when dynamically masking out the visual objects. However, we notice in surprise that when all visual objects mentioned in the instruction are masked out, the agents can still reach a success rate higher than 44\% on R2R and higher than 32\% on RxR-en.
This contradicts the previous assumption and casts doubt on the reliability of the navigation agents' vision-language alignment.

Comparing ``dynamic mask'' with the ``mask tokens'' setting, we notice that the visual object ablation has much smaller impact on navigation performance than the  text object ablations, which suggests that {\uline{current models have unbalanced attention on vision and text for the VLN task}}. Recent studies on pre-trained vision-and-language models~\citep{Frank2021VisionandLanguageOV} reveal that such asymmetry is also witnessed in other multimodal tasks. Future studies may follow the line of constructing a more balanced VLN agent.

%======================================================================================
\section{Conclusion}
In this paper, we inspect how the navigation agents understand the multimodal information by conducting ablation diagnostics input data. 
We find out that indoor navigation agents refer to both object tokens and direction tokens in the instruction when making decisions. In contrast, outdoor navigation agents heavily rely on direction tokens and poorly understand the object tokens. 
When it comes to vision-and-language alignments, we witness unbalanced attention on text and vision, and doubt the reliability of cross-modal alignments.
We hope this work encourages more investigation and research into understanding neural VLN agents' black-box and improves the task setups and navigation agents' capacity for future studies. 

\section*{Acknowledgement}
We would like to show our gratitude to the anonymous reviewers for their thought-provoking comments.
We would like to thank the Robert N. Noyce Trust for their generous gift to the University of California via the Noyce Initiative.
The UCSB authors were also sponsored by the U.S. Army Research Office, and this work was accomplished under Contract Number W911NF19-D-0001 for the Institute for Collaborative Biotechnologies.
The views and conclusions contained in this research are those of the authors and should not be interpreted as representing the sponsors or the U.S. government's official policy, expressed or inferred. Regardless of any copyright notation herein, the United States government is authorized to reproduce and distribute reprints for government purposes.

\bibliography{anthology,custom}

\begin{thebibliography}{70}
\expandafter\ifx\csname natexlab\endcsname\relax\def\natexlab#1{#1}\fi

\bibitem[{Anderson et~al.(1991)Anderson, Bader, Bard, Boyle, Doherty, Garrod,
  Isard, Kowtko, McAllister, Miller, Sotillo, Thompson, and
  Weinert}]{Anderson1991TheHM}
Anne~H. Anderson, Miles Bader, Ellen~Gurman Bard, Elizabeth Boyle, Gwyneth
  Doherty, Simon Garrod, Stephen~D. Isard, Jacqueline~C. Kowtko, Jan
  McAllister, Jim Miller, Catherine Sotillo, Henry~S. Thompson, and Regina
  Weinert. 1991.
\newblock The hcrc map task corpus.
\newblock \emph{Language and Speech}, 34:351 -- 366.

\bibitem[{Anderson et~al.(2018)Anderson, Wu, Teney, Bruce, Johnson,
  S{\"{u}}nderhauf, Reid, Gould, and van~den
  Hengel}]{DBLP:conf/cvpr/AndersonWTB0S0G18}
Peter Anderson, Qi~Wu, Damien Teney, Jake Bruce, Mark Johnson, Niko
  S{\"{u}}nderhauf, Ian~D. Reid, Stephen Gould, and Anton van~den Hengel. 2018.
\newblock \href {https://doi.org/10.1109/CVPR.2018.00387} {Vision-and-language
  navigation: Interpreting visually-grounded navigation instructions in real
  environments}.
\newblock In \emph{2018 {IEEE} Conference on Computer Vision and Pattern
  Recognition, {CVPR} 2018, Salt Lake City, UT, USA, June 18-22, 2018}, pages
  3674--3683. {IEEE} Computer Society.

\bibitem[{Andreas and Klein(2015)}]{andreas-klein-2015-alignment}
Jacob Andreas and Dan Klein. 2015.
\newblock \href {https://doi.org/10.18653/v1/D15-1138} {Alignment-based
  compositional semantics for instruction following}.
\newblock In \emph{Proceedings of the 2015 Conference on Empirical Methods in
  Natural Language Processing}, pages 1165--1174, Lisbon, Portugal. Association
  for Computational Linguistics.

\bibitem[{Antoniol et~al.(2011)Antoniol, Cattoni, and
  Cettolo}]{Antoniol2011RobustSU}
Giuliano Antoniol, Roldano Cattoni, and Mauro Cettolo. 2011.
\newblock Robust speech understanding for robot telecontrol.

\bibitem[{Artzi and Zettlemoyer(2013)}]{Artzi2013WeaklySL}
Yoav Artzi and Luke Zettlemoyer. 2013.
\newblock Weakly supervised learning of semantic parsers for mapping
  instructions to actions.
\newblock \emph{Transactions of the Association for Computational Linguistics},
  1:49--62.

\bibitem[{Bisk et~al.(2016{\natexlab{a}})Bisk, Yuret, and
  Marcu}]{Bisk2016NaturalLC}
Yonatan Bisk, Deniz Yuret, and Daniel Marcu. 2016{\natexlab{a}}.
\newblock Natural language communication with robots.
\newblock In \emph{NAACL}.

\bibitem[{Bisk et~al.(2016{\natexlab{b}})Bisk, Yuret, and
  Marcu}]{bisk-etal-2016-natural}
Yonatan Bisk, Deniz Yuret, and Daniel Marcu. 2016{\natexlab{b}}.
\newblock \href {https://doi.org/10.18653/v1/N16-1089} {Natural language
  communication with robots}.
\newblock In \emph{Proceedings of the 2016 Conference of the North {A}merican
  Chapter of the Association for Computational Linguistics: Human Language
  Technologies}, pages 751--761, San Diego, California. Association for
  Computational Linguistics.

\bibitem[{Branavan et~al.(2009)Branavan, Chen, Zettlemoyer, and
  Barzilay}]{branavan-etal-2009-reinforcement}
S.R.K. Branavan, Harr Chen, Luke Zettlemoyer, and Regina Barzilay. 2009.
\newblock \href {https://aclanthology.org/P09-1010} {Reinforcement learning for
  mapping instructions to actions}.
\newblock In \emph{Proceedings of the Joint Conference of the 47th Annual
  Meeting of the {ACL} and the 4th International Joint Conference on Natural
  Language Processing of the {AFNLP}}, pages 82--90, Suntec, Singapore.
  Association for Computational Linguistics.

\bibitem[{Branavan et~al.(2010)Branavan, Zettlemoyer, and
  Barzilay}]{branavan-etal-2010-reading}
S.R.K. Branavan, Luke Zettlemoyer, and Regina Barzilay. 2010.
\newblock \href {https://aclanthology.org/P10-1129} {Reading between the lines:
  Learning to map high-level instructions to commands}.
\newblock In \emph{Proceedings of the 48th Annual Meeting of the Association
  for Computational Linguistics}, pages 1268--1277, Uppsala, Sweden.
  Association for Computational Linguistics.

\bibitem[{Chang et~al.(2017)Chang, Dai, Funkhouser, Halber, Nie{\ss}ner, Savva,
  Song, Zeng, and Zhang}]{DBLP:conf/3dim/ChangDFHNSSZZ17}
Angel~X. Chang, Angela Dai, Thomas~A. Funkhouser, Maciej Halber, Matthias
  Nie{\ss}ner, Manolis Savva, Shuran Song, Andy Zeng, and Yinda Zhang. 2017.
\newblock \href {https://doi.org/10.1109/3DV.2017.00081} {Matterport3d:
  Learning from {RGB-D} data in indoor environments}.
\newblock In \emph{2017 International Conference on 3D Vision, 3DV 2017,
  Qingdao, China, October 10-12, 2017}, pages 667--676. {IEEE} Computer
  Society.

\bibitem[{Chen and Mooney(2011)}]{Chen2011LearningTI}
David~L. Chen and Raymond~J. Mooney. 2011.
\newblock Learning to interpret natural language navigation instructions from
  observations.
\newblock In \emph{AAAI 2011}.

\bibitem[{Chen et~al.(2019)Chen, Suhr, Misra, Snavely, and
  Artzi}]{DBLP:conf/cvpr/ChenSMSA19}
Howard Chen, Alane Suhr, Dipendra Misra, Noah Snavely, and Yoav Artzi. 2019.
\newblock \href {https://doi.org/10.1109/CVPR.2019.01282} {{TOUCHDOWN:} natural
  language navigation and spatial reasoning in visual street environments}.
\newblock In \emph{{IEEE} Conference on Computer Vision and Pattern
  Recognition, {CVPR} 2019, Long Beach, CA, USA, June 16-20, 2019}, pages
  12538--12547. Computer Vision Foundation / {IEEE}.

\bibitem[{Chen et~al.(2021)Chen, Guhur, Schmid, and Laptev}]{chen2021hamt}
Shizhe Chen, Pierre-Louis Guhur, Cordelia Schmid, and Ivan Laptev. 2021.
\newblock History aware multimodal transformer for vision-and-language
  navigation.
\newblock In \emph{NeurIPS}.

\bibitem[{Dan et~al.(2021)Dan, Zhou, and Roth}]{dan-etal-2021-generalization}
Soham Dan, Michael Zhou, and Dan Roth. 2021.
\newblock \href {https://doi.org/10.18653/v1/2021.naacl-main.76}
  {Generalization in instruction following systems}.
\newblock In \emph{Proceedings of the 2021 Conference of the North American
  Chapter of the Association for Computational Linguistics: Human Language
  Technologies}, pages 976--981, Online. Association for Computational
  Linguistics.

\bibitem[{Fernandes et~al.(2021)Fernandes, Yin, Neubig, and
  Martins}]{fernandes-etal-2021-measuring}
Patrick Fernandes, Kayo Yin, Graham Neubig, and Andr{\'e} F.~T. Martins. 2021.
\newblock \href {https://doi.org/10.18653/v1/2021.acl-long.505} {Measuring and
  increasing context usage in context-aware machine translation}.
\newblock In \emph{Proceedings of the 59th Annual Meeting of the Association
  for Computational Linguistics and the 11th International Joint Conference on
  Natural Language Processing (Volume 1: Long Papers)}, pages 6467--6478,
  Online. Association for Computational Linguistics.

\bibitem[{Frank et~al.(2021)Frank, Bugliarello, and
  Elliott}]{Frank2021VisionandLanguageOV}
Stella Frank, Emanuele Bugliarello, and Desmond Elliott. 2021.
\newblock Vision-and-language or vision-for-language? on cross-modal influence
  in multimodal transformers.
\newblock In \emph{EMNLP}.

\bibitem[{Fried et~al.(2018)Fried, Hu, Cirik, Rohrbach, Andreas, Morency,
  Berg{-}Kirkpatrick, Saenko, Klein, and
  Darrell}]{DBLP:conf/nips/FriedHCRAMBSKD18}
Daniel Fried, Ronghang Hu, Volkan Cirik, Anna Rohrbach, Jacob Andreas,
  Louis{-}Philippe Morency, Taylor Berg{-}Kirkpatrick, Kate Saenko, Dan Klein,
  and Trevor Darrell. 2018.
\newblock \href
  {https://proceedings.neurips.cc/paper/2018/hash/6a81681a7af700c6385d36577ebec359-Abstract.html}
  {Speaker-follower models for vision-and-language navigation}.
\newblock In \emph{Advances in Neural Information Processing Systems 31: Annual
  Conference on Neural Information Processing Systems 2018, NeurIPS 2018,
  December 3-8, 2018, Montr{\'{e}}al, Canada}, pages 3318--3329.

\bibitem[{Guadarrama et~al.(2013)Guadarrama, Riano, Golland, Goehring, Jia,
  Klein, Abbeel, and Darrell}]{Guadarrama2013GroundingSR}
Sergio Guadarrama, Lorenzo Riano, David~Hamilton Golland, Daniel Goehring,
  Yangqing Jia, Dan Klein, P.~Abbeel, and Trevor Darrell. 2013.
\newblock Grounding spatial relations for human-robot interaction.
\newblock \emph{2013 IEEE/RSJ International Conference on Intelligent Robots
  and Systems}, pages 1640--1647.

\bibitem[{Hao et~al.(2020)Hao, Li, Li, Carin, and
  Gao}]{DBLP:conf/cvpr/HaoLLCG20}
Weituo Hao, Chunyuan Li, Xiujun Li, Lawrence Carin, and Jianfeng Gao. 2020.
\newblock \href {https://doi.org/10.1109/CVPR42600.2020.01315} {Towards
  learning a generic agent for vision-and-language navigation via
  pre-training}.
\newblock In \emph{2020 {IEEE/CVF} Conference on Computer Vision and Pattern
  Recognition, {CVPR} 2020, Seattle, WA, USA, June 13-19, 2020}, pages
  13134--13143. {IEEE}.

\bibitem[{He et~al.(2016)He, Zhang, Ren, and Sun}]{DBLP:conf/cvpr/HeZRS16}
Kaiming He, Xiangyu Zhang, Shaoqing Ren, and Jian Sun. 2016.
\newblock \href {https://doi.org/10.1109/CVPR.2016.90} {Deep residual learning
  for image recognition}.
\newblock In \emph{2016 {IEEE} Conference on Computer Vision and Pattern
  Recognition, {CVPR} 2016, Las Vegas, NV, USA, June 27-30, 2016}, pages
  770--778. {IEEE} Computer Society.

\bibitem[{Hendricks et~al.(2016)Hendricks, Akata, Rohrbach, Donahue, Schiele,
  and Darrell}]{DBLP:conf/eccv/HendricksARDSD16}
Lisa~Anne Hendricks, Zeynep Akata, Marcus Rohrbach, Jeff Donahue, Bernt
  Schiele, and Trevor Darrell. 2016.
\newblock \href {https://doi.org/10.1007/978-3-319-46493-0\_1} {Generating
  visual explanations}.
\newblock In \emph{Computer Vision - {ECCV} 2016 - 14th European Conference,
  Amsterdam, The Netherlands, October 11-14, 2016, Proceedings, Part {IV}},
  volume 9908 of \emph{Lecture Notes in Computer Science}, pages 3--19.
  Springer.

\bibitem[{Hong et~al.(2020{\natexlab{a}})Hong, Opazo, Qi, Wu, and
  Gould}]{DBLP:conf/nips/HongOQ0G20}
Yicong Hong, Cristian~Rodriguez Opazo, Yuankai Qi, Qi~Wu, and Stephen Gould.
  2020{\natexlab{a}}.
\newblock \href
  {https://proceedings.neurips.cc/paper/2020/hash/56dc0997d871e9177069bb472574eb29-Abstract.html}
  {Language and visual entity relationship graph for agent navigation}.
\newblock In \emph{Advances in Neural Information Processing Systems 33: Annual
  Conference on Neural Information Processing Systems 2020, NeurIPS 2020,
  December 6-12, 2020, virtual}.

\bibitem[{Hong et~al.(2020{\natexlab{b}})Hong, Wu, Qi, Opazo, and
  Gould}]{DBLP:journals/corr/abs-2011-13922}
Yicong Hong, Qi~Wu, Yuankai Qi, Cristian~Rodriguez Opazo, and Stephen Gould.
  2020{\natexlab{b}}.
\newblock \href {http://arxiv.org/abs/2011.13922} {A recurrent
  vision-and-language {BERT} for navigation}.
\newblock \emph{CoRR}, abs/2011.13922.

\bibitem[{Huang et~al.(2019)Huang, Jain, Mehta, Ku, Magalh{\~{a}}es, Baldridge,
  and Ie}]{DBLP:conf/iccv/HuangJMKMBI19}
Haoshuo Huang, Vihan Jain, Harsh Mehta, Alexander Ku, Gabriel Magalh{\~{a}}es,
  Jason Baldridge, and Eugene Ie. 2019.
\newblock \href {https://doi.org/10.1109/ICCV.2019.00750} {Transferable
  representation learning in vision-and-language navigation}.
\newblock In \emph{2019 {IEEE/CVF} International Conference on Computer Vision,
  {ICCV} 2019, Seoul, Korea (South), October 27 - November 2, 2019}, pages
  7403--7412. {IEEE}.

\bibitem[{Jain et~al.(2019{\natexlab{a}})Jain, Magalh{\~{a}}es, Ku, Vaswani,
  Ie, and Baldridge}]{DBLP:conf/acl/JainMKVIB19}
Vihan Jain, Gabriel Magalh{\~{a}}es, Alexander Ku, Ashish Vaswani, Eugene Ie,
  and Jason Baldridge. 2019{\natexlab{a}}.
\newblock \href {https://doi.org/10.18653/v1/p19-1181} {Stay on the path:
  Instruction fidelity in vision-and-language navigation}.
\newblock In \emph{Proceedings of the 57th Conference of the Association for
  Computational Linguistics, {ACL} 2019, Florence, Italy, July 28- August 2,
  2019, Volume 1: Long Papers}, pages 1862--1872. Association for Computational
  Linguistics.

\bibitem[{Jain et~al.(2019{\natexlab{b}})Jain, Magalh{\~a}es, Ku, Vaswani, Ie,
  and Baldridge}]{Jain2019StayOT}
Vihan Jain, Gabriel Magalh{\~a}es, Alexander Ku, Ashish Vaswani, Eugene Ie, and
  Jason Baldridge. 2019{\natexlab{b}}.
\newblock Stay on the path: Instruction fidelity in vision-and-language
  navigation.
\newblock \emph{ArXiv}, abs/1905.12255.

\bibitem[{Ke et~al.(2019)Ke, Li, Bisk, Holtzman, Gan, Liu, Gao, Choi, and
  Srinivasa}]{DBLP:conf/cvpr/KeLBHGLGCS19}
Liyiming Ke, Xiujun Li, Yonatan Bisk, Ari Holtzman, Zhe Gan, Jingjing Liu,
  Jianfeng Gao, Yejin Choi, and Siddhartha~S. Srinivasa. 2019.
\newblock \href {https://doi.org/10.1109/CVPR.2019.00690} {Tactical rewind:
  Self-correction via backtracking in vision-and-language navigation}.
\newblock In \emph{{IEEE} Conference on Computer Vision and Pattern
  Recognition, {CVPR} 2019, Long Beach, CA, USA, June 16-20, 2019}, pages
  6741--6749. Computer Vision Foundation / {IEEE}.

\bibitem[{Klingspor et~al.(1997)Klingspor, Demiris, and
  Kaiser}]{Klingspor97human-robot-communicationand}
Volker Klingspor, John Demiris, and Michael Kaiser. 1997.
\newblock Human-robot-communication and machine learning.
\newblock \emph{APPLIED ARTIFICIAL INTELLIGENCE JOURNAL}, 11(11):719--746.

\bibitem[{Kollar et~al.(2013)Kollar, Krishnamurthy, and
  Strimel}]{Kollar2013TowardIG}
Thomas Kollar, Jayant Krishnamurthy, and Grant~P. Strimel. 2013.
\newblock Toward interactive grounded language acqusition.
\newblock In \emph{Robotics: Science and Systems}.

\bibitem[{Ku et~al.(2020{\natexlab{a}})Ku, Anderson, Patel, Ie, and
  Baldridge}]{DBLP:conf/emnlp/KuAPIB20}
Alexander Ku, Peter Anderson, Roma Patel, Eugene Ie, and Jason Baldridge.
  2020{\natexlab{a}}.
\newblock \href {https://doi.org/10.18653/v1/2020.emnlp-main.356}
  {Room-across-room: Multilingual vision-and-language navigation with dense
  spatiotemporal grounding}.
\newblock In \emph{Proceedings of the 2020 Conference on Empirical Methods in
  Natural Language Processing, {EMNLP} 2020, Online, November 16-20, 2020},
  pages 4392--4412. Association for Computational Linguistics.

\bibitem[{Ku et~al.(2020{\natexlab{b}})Ku, Anderson, Patel, Ie, and
  Baldridge}]{rxr}
Alexander Ku, Peter Anderson, Roma Patel, Eugene Ie, and Jason Baldridge.
  2020{\natexlab{b}}.
\newblock {Room-Across-Room}: Multilingual vision-and-language navigation with
  dense spatiotemporal grounding.
\newblock In \emph{Conference on Empirical Methods for Natural Language
  Processing (EMNLP)}.

\bibitem[{Li et~al.(2018)Li, Fu, Yu, Mei, and Luo}]{DBLP:conf/emnlp/LiFYML18}
Qing Li, Jianlong Fu, Dongfei Yu, Tao Mei, and Jiebo Luo. 2018.
\newblock \href {https://doi.org/10.18653/v1/d18-1164} {Tell-and-answer:
  Towards explainable visual question answering using attributes and captions}.
\newblock In \emph{Proceedings of the 2018 Conference on Empirical Methods in
  Natural Language Processing, Brussels, Belgium, October 31 - November 4,
  2018}, pages 1338--1346. Association for Computational Linguistics.

\bibitem[{Li et~al.(2019)Li, Li, Xia, Bisk, Celikyilmaz, Gao, Smith, and
  Choi}]{DBLP:conf/emnlp/LiLXBCGSC19}
Xiujun Li, Chunyuan Li, Qiaolin Xia, Yonatan Bisk, Asli Celikyilmaz, Jianfeng
  Gao, Noah~A. Smith, and Yejin Choi. 2019.
\newblock \href {https://doi.org/10.18653/v1/D19-1159} {Robust navigation with
  language pretraining and stochastic sampling}.
\newblock In \emph{Proceedings of the 2019 Conference on Empirical Methods in
  Natural Language Processing and the 9th International Joint Conference on
  Natural Language Processing, {EMNLP-IJCNLP} 2019, Hong Kong, China, November
  3-7, 2019}, pages 1494--1499. Association for Computational Linguistics.

\bibitem[{Lu et~al.(2016)Lu, Yang, Batra, and Parikh}]{DBLP:conf/nips/LuYBP16}
Jiasen Lu, Jianwei Yang, Dhruv Batra, and Devi Parikh. 2016.
\newblock \href
  {https://proceedings.neurips.cc/paper/2016/hash/9dcb88e0137649590b755372b040afad-Abstract.html}
  {Hierarchical question-image co-attention for visual question answering}.
\newblock In \emph{Advances in Neural Information Processing Systems 29: Annual
  Conference on Neural Information Processing Systems 2016, December 5-10,
  2016, Barcelona, Spain}, pages 289--297.

\bibitem[{Ma et~al.(2019{\natexlab{a}})Ma, Lu, Wu, AlRegib, Kira, Socher, and
  Xiong}]{DBLP:conf/iclr/MaLWAKSX19}
Chih{-}Yao Ma, Jiasen Lu, Zuxuan Wu, Ghassan AlRegib, Zsolt Kira, Richard
  Socher, and Caiming Xiong. 2019{\natexlab{a}}.
\newblock \href {https://openreview.net/forum?id=r1GAsjC5Fm} {Self-monitoring
  navigation agent via auxiliary progress estimation}.
\newblock In \emph{7th International Conference on Learning Representations,
  {ICLR} 2019, New Orleans, LA, USA, May 6-9, 2019}. OpenReview.net.

\bibitem[{Ma et~al.(2019{\natexlab{b}})Ma, Wu, AlRegib, Xiong, and
  Kira}]{DBLP:conf/cvpr/MaWAXK19}
Chih{-}Yao Ma, Zuxuan Wu, Ghassan AlRegib, Caiming Xiong, and Zsolt Kira.
  2019{\natexlab{b}}.
\newblock \href {https://doi.org/10.1109/CVPR.2019.00689} {The regretful agent:
  Heuristic-aided navigation through progress estimation}.
\newblock In \emph{{IEEE} Conference on Computer Vision and Pattern
  Recognition, {CVPR} 2019, Long Beach, CA, USA, June 16-20, 2019}, pages
  6732--6740. Computer Vision Foundation / {IEEE}.

\bibitem[{Matuszek et~al.(2014)Matuszek, Bo, Zettlemoyer, and
  Fox}]{Matuszek2014LearningFU}
Cynthia Matuszek, Liefeng Bo, Luke Zettlemoyer, and Dieter Fox. 2014.
\newblock Learning from unscripted deictic gesture and language for human-robot
  interactions.
\newblock In \emph{AAAI}.

\bibitem[{Mehta et~al.(2020)Mehta, Artzi, Baldridge, Ie, and
  Mirowski}]{DBLP:journals/corr/abs-2001-03671}
Harsh Mehta, Yoav Artzi, Jason Baldridge, Eugene Ie, and Piotr Mirowski. 2020.
\newblock \href {http://arxiv.org/abs/2001.03671} {Retouchdown: Adding
  touchdown to streetlearn as a shareable resource for language grounding tasks
  in street view}.
\newblock \emph{CoRR}, abs/2001.03671.

\bibitem[{Mei et~al.(2016)Mei, Bansal, and Walter}]{Mei2016ListenAA}
Hongyuan Mei, Mohit Bansal, and Matthew~R. Walter. 2016.
\newblock Listen, attend, and walk: Neural mapping of navigational instructions
  to action sequences.
\newblock In \emph{AAAI}.

\bibitem[{Mirowski et~al.(2018)Mirowski, Grimes, Malinowski, Hermann, Anderson,
  Teplyashin, Simonyan, Kavukcuoglu, Zisserman, and
  Hadsell}]{DBLP:conf/nips/MirowskiGMHATSK18}
Piotr Mirowski, Matthew~Koichi Grimes, Mateusz Malinowski, Karl~Moritz Hermann,
  Keith Anderson, Denis Teplyashin, Karen Simonyan, Koray Kavukcuoglu, Andrew
  Zisserman, and Raia Hadsell. 2018.
\newblock \href
  {https://proceedings.neurips.cc/paper/2018/hash/e034fb6b66aacc1d48f445ddfb08da98-Abstract.html}
  {Learning to navigate in cities without a map}.
\newblock In \emph{Advances in Neural Information Processing Systems 31: Annual
  Conference on Neural Information Processing Systems 2018, NeurIPS 2018,
  December 3-8, 2018, Montr{\'{e}}al, Canada}, pages 2424--2435.

\bibitem[{Misra et~al.(2018)Misra, Bennett, Blukis, Niklasson, Shatkhin, and
  Artzi}]{Misra2018MappingIT}
Dipendra Misra, Andrew Bennett, Valts Blukis, Eyvind Niklasson, Max Shatkhin,
  and Yoav Artzi. 2018.
\newblock Mapping instructions to actions in 3d environments with visual goal
  prediction.
\newblock In \emph{EMNLP}.

\bibitem[{Misra et~al.(2017)Misra, Langford, and Artzi}]{Misra2017MappingIA}
Dipendra~Kumar Misra, John Langford, and Yoav Artzi. 2017.
\newblock Mapping instructions and visual observations to actions with
  reinforcement learning.
\newblock In \emph{EMNLP}.

\bibitem[{Nguyen et~al.(2019)Nguyen, Dey, Brockett, and
  Dolan}]{DBLP:conf/cvpr/NguyenDBD19}
Khanh Nguyen, Debadeepta Dey, Chris Brockett, and Bill Dolan. 2019.
\newblock \href {https://doi.org/10.1109/CVPR.2019.01281} {Vision-based
  navigation with language-based assistance via imitation learning with
  indirect intervention}.
\newblock In \emph{{IEEE} Conference on Computer Vision and Pattern
  Recognition, {CVPR} 2019, Long Beach, CA, USA, June 16-20, 2019}, pages
  12527--12537. Computer Vision Foundation / {IEEE}.

\bibitem[{Nguyen and III(2019)}]{DBLP:conf/emnlp/NguyenD19}
Khanh Nguyen and Hal~Daum{\'{e}} III. 2019.
\newblock \href {https://doi.org/10.18653/v1/D19-1063} {Help, anna! visual
  navigation with natural multimodal assistance via retrospective
  curiosity-encouraging imitation learning}.
\newblock In \emph{Proceedings of the 2019 Conference on Empirical Methods in
  Natural Language Processing and the 9th International Joint Conference on
  Natural Language Processing, {EMNLP-IJCNLP} 2019, Hong Kong, China, November
  3-7, 2019}, pages 684--695. Association for Computational Linguistics.

\bibitem[{O'Connor and Andreas(2021)}]{OConnor2021WhatCF}
Joe O'Connor and Jacob Andreas. 2021.
\newblock What context features can transformer language models use?
\newblock In \emph{ACL/IJCNLP}.

\bibitem[{Park et~al.(2018)Park, Hendricks, Akata, Rohrbach, Schiele, Darrell,
  and Rohrbach}]{DBLP:conf/cvpr/ParkHARSDR18}
Dong~Huk Park, Lisa~Anne Hendricks, Zeynep Akata, Anna Rohrbach, Bernt Schiele,
  Trevor Darrell, and Marcus Rohrbach. 2018.
\newblock \href {https://doi.org/10.1109/CVPR.2018.00915} {Multimodal
  explanations: Justifying decisions and pointing to the evidence}.
\newblock In \emph{2018 {IEEE} Conference on Computer Vision and Pattern
  Recognition, {CVPR} 2018, Salt Lake City, UT, USA, June 18-22, 2018}, pages
  8779--8788. {IEEE} Computer Society.

\bibitem[{Qi et~al.(2020{\natexlab{a}})Qi, Zhang, Zhang, Bolton, and
  Manning}]{DBLP:conf/acl/QiZZBM20}
Peng Qi, Yuhao Zhang, Yuhui Zhang, Jason Bolton, and Christopher~D. Manning.
  2020{\natexlab{a}}.
\newblock \href {https://doi.org/10.18653/v1/2020.acl-demos.14} {Stanza: {A}
  python natural language processing toolkit for many human languages}.
\newblock In \emph{Proceedings of the 58th Annual Meeting of the Association
  for Computational Linguistics: System Demonstrations, {ACL} 2020, Online,
  July 5-10, 2020}, pages 101--108. Association for Computational Linguistics.

\bibitem[{Qi et~al.(2020{\natexlab{b}})Qi, Pan, Zhang, van~den Hengel, and
  Wu}]{DBLP:conf/eccv/QiPZHW20}
Yuankai Qi, Zizheng Pan, Shengping Zhang, Anton van~den Hengel, and Qi~Wu.
  2020{\natexlab{b}}.
\newblock \href {https://doi.org/10.1007/978-3-030-58607-2\_18}
  {Object-and-action aware model for visual language navigation}.
\newblock In \emph{Computer Vision - {ECCV} 2020 - 16th European Conference,
  Glasgow, UK, August 23-28, 2020, Proceedings, Part {X}}, volume 12355 of
  \emph{Lecture Notes in Computer Science}, pages 303--317. Springer.

\bibitem[{Qi et~al.(2020{\natexlab{c}})Qi, Wu, Anderson, Wang, Wang, Shen, and
  van~den Hengel}]{DBLP:conf/cvpr/QiW0WWSH20}
Yuankai Qi, Qi~Wu, Peter Anderson, Xin Wang, William~Yang Wang, Chunhua Shen,
  and Anton van~den Hengel. 2020{\natexlab{c}}.
\newblock \href {https://doi.org/10.1109/CVPR42600.2020.01000} {{REVERIE:}
  remote embodied visual referring expression in real indoor environments}.
\newblock In \emph{2020 {IEEE/CVF} Conference on Computer Vision and Pattern
  Recognition, {CVPR} 2020, Seattle, WA, USA, June 13-19, 2020}, pages
  9979--9988. {IEEE}.

\bibitem[{Radford et~al.(2021)Radford, Kim, Hallacy, Ramesh, Goh, Agarwal,
  Sastry, Askell, Mishkin, Clark, Krueger, and
  Sutskever}]{Radford2021LearningTV}
Alec Radford, Jong~Wook Kim, Chris Hallacy, Aditya Ramesh, Gabriel Goh,
  Sandhini Agarwal, Girish Sastry, Amanda Askell, Pamela Mishkin, Jack Clark,
  Gretchen Krueger, and Ilya Sutskever. 2021.
\newblock Learning transferable visual models from natural language
  supervision.
\newblock In \emph{ICML}.

\bibitem[{Roy(2002)}]{Roy2002LearningVG}
Deb~K. Roy. 2002.
\newblock Learning visually grounded words and syntax for a scene description
  task.
\newblock \emph{Comput. Speech Lang.}, 16:353--385.

\bibitem[{Selvaraju et~al.(2017)Selvaraju, Cogswell, Das, Vedantam, Parikh, and
  Batra}]{DBLP:conf/iccv/SelvarajuCDVPB17}
Ramprasaath~R. Selvaraju, Michael Cogswell, Abhishek Das, Ramakrishna Vedantam,
  Devi Parikh, and Dhruv Batra. 2017.
\newblock \href {https://doi.org/10.1109/ICCV.2017.74} {Grad-cam: Visual
  explanations from deep networks via gradient-based localization}.
\newblock In \emph{{IEEE} International Conference on Computer Vision, {ICCV}
  2017, Venice, Italy, October 22-29, 2017}, pages 618--626. {IEEE} Computer
  Society.

\bibitem[{Shen et~al.(2021)Shen, Li, Tan, Bansal, Rohrbach, Chang, Yao, and
  Keutzer}]{Shen2021HowMC}
Sheng Shen, Liunian~Harold Li, Hao Tan, Mohit Bansal, Anna Rohrbach, Kai-Wei
  Chang, Zhewei Yao, and Kurt Keutzer. 2021.
\newblock How much can clip benefit vision-and-language tasks?
\newblock \emph{ArXiv}, abs/2107.06383.

\bibitem[{Steels and Vogt(1997)}]{Steels1997GroundingAL}
Luc~L. Steels and Paul Vogt. 1997.
\newblock Grounding adaptive language games in robotic agents.

\bibitem[{Szegedy et~al.(2017)Szegedy, Ioffe, Vanhoucke, and
  Alemi}]{Szegedy2017Inceptionv4IA}
Christian Szegedy, Sergey Ioffe, Vincent Vanhoucke, and Alexander~Amir Alemi.
  2017.
\newblock Inception-v4, inception-resnet and the impact of residual connections
  on learning.
\newblock In \emph{AAAI}.

\bibitem[{Tan and Bansal(2018)}]{Tan2018SourceTargetIM}
Hao Tan and Mohit Bansal. 2018.
\newblock Source-target inference models for spatial instruction understanding.
\newblock In \emph{AAAI}.

\bibitem[{Tan et~al.(2019)Tan, Yu, and Bansal}]{DBLP:conf/naacl/TanYB19}
Hao Tan, Licheng Yu, and Mohit Bansal. 2019.
\newblock \href {https://doi.org/10.18653/v1/n19-1268} {Learning to navigate
  unseen environments: Back translation with environmental dropout}.
\newblock In \emph{Proceedings of the 2019 Conference of the North American
  Chapter of the Association for Computational Linguistics: Human Language
  Technologies, {NAACL-HLT} 2019, Minneapolis, MN, USA, June 2-7, 2019, Volume
  1 (Long and Short Papers)}, pages 2610--2621. Association for Computational
  Linguistics.

\bibitem[{Thomason et~al.(2019)Thomason, Gordon, and
  Bisk}]{Thomason2019ShiftingTB}
Jesse Thomason, Daniel Gordon, and Yonatan Bisk. 2019.
\newblock Shifting the baseline: Single modality performance on visual
  navigation \& qa.
\newblock In \emph{NAACL}.

\bibitem[{Vogel and Jurafsky(2010)}]{vogel-jurafsky-2010-learning}
Adam Vogel and Daniel Jurafsky. 2010.
\newblock \href {https://aclanthology.org/P10-1083} {Learning to follow
  navigational directions}.
\newblock In \emph{Proceedings of the 48th Annual Meeting of the Association
  for Computational Linguistics}, pages 806--814, Uppsala, Sweden. Association
  for Computational Linguistics.

\bibitem[{Wang et~al.(2020{\natexlab{a}})Wang, Wu, and
  Shen}]{DBLP:conf/eccv/WangWS20}
Hu~Wang, Qi~Wu, and Chunhua Shen. 2020{\natexlab{a}}.
\newblock \href {https://doi.org/10.1007/978-3-030-58545-7\_8} {Soft expert
  reward learning for vision-and-language navigation}.
\newblock In \emph{Computer Vision - {ECCV} 2020 - 16th European Conference,
  Glasgow, UK, August 23-28, 2020, Proceedings, Part {IX}}, volume 12354 of
  \emph{Lecture Notes in Computer Science}, pages 126--141. Springer.

\bibitem[{Wang et~al.(2019)Wang, Huang, Celikyilmaz, Gao, Shen, Wang, Wang, and
  Zhang}]{DBLP:conf/cvpr/WangHcGSWWZ19}
Xin Wang, Qiuyuan Huang, Asli Celikyilmaz, Jianfeng Gao, Dinghan Shen,
  Yuan{-}Fang Wang, William~Yang Wang, and Lei Zhang. 2019.
\newblock \href {https://doi.org/10.1109/CVPR.2019.00679} {Reinforced
  cross-modal matching and self-supervised imitation learning for
  vision-language navigation}.
\newblock In \emph{{IEEE} Conference on Computer Vision and Pattern
  Recognition, {CVPR} 2019, Long Beach, CA, USA, June 16-20, 2019}, pages
  6629--6638. Computer Vision Foundation / {IEEE}.

\bibitem[{Wang et~al.(2018)Wang, Xiong, Wang, and
  Wang}]{DBLP:conf/eccv/WangXWW18}
Xin Wang, Wenhan Xiong, Hongmin Wang, and William~Yang Wang. 2018.
\newblock \href {https://doi.org/10.1007/978-3-030-01270-0\_3} {Look before you
  leap: Bridging model-free and model-based reinforcement learning for
  planned-ahead vision-and-language navigation}.
\newblock In \emph{Computer Vision - {ECCV} 2018 - 15th European Conference,
  Munich, Germany, September 8-14, 2018, Proceedings, Part {XVI}}, volume 11220
  of \emph{Lecture Notes in Computer Science}, pages 38--55. Springer.

\bibitem[{Wang et~al.(2020{\natexlab{b}})Wang, Jain, Ie, Wang, Kozareva, and
  Ravi}]{DBLP:conf/eccv/WangJIWKR20}
Xin~Eric Wang, Vihan Jain, Eugene Ie, William~Yang Wang, Zornitsa Kozareva, and
  Sujith Ravi. 2020{\natexlab{b}}.
\newblock \href {https://doi.org/10.1007/978-3-030-58586-0\_25}
  {Environment-agnostic multitask learning for natural language grounded
  navigation}.
\newblock In \emph{Computer Vision - {ECCV} 2020 - 16th European Conference,
  Glasgow, UK, August 23-28, 2020, Proceedings, Part {XXIV}}, volume 12369 of
  \emph{Lecture Notes in Computer Science}, pages 413--430. Springer.

\bibitem[{Winograd(1971)}]{Winograd1971ProceduresAA}
Terry Winograd. 1971.
\newblock Procedures as a representation for data in a computer program for
  understanding natural language.

\bibitem[{Wu and Mooney(2019)}]{wu-mooney-2019-faithful}
Jialin Wu and Raymond Mooney. 2019.
\newblock \href {https://doi.org/10.18653/v1/W19-4812} {Faithful multimodal
  explanation for visual question answering}.
\newblock In \emph{Proceedings of the 2019 ACL Workshop BlackboxNLP: Analyzing
  and Interpreting Neural Networks for NLP}, pages 103--112, Florence, Italy.
  Association for Computational Linguistics.

\bibitem[{Xia et~al.(2020)Xia, Li, Li, Bisk, Sui, Gao, Choi, and
  Smith}]{DBLP:journals/corr/abs-2003-00857}
Qiaolin Xia, Xiujun Li, Chunyuan Li, Yonatan Bisk, Zhifang Sui, Jianfeng Gao,
  Yejin Choi, and Noah~A. Smith. 2020.
\newblock \href {http://arxiv.org/abs/2003.00857} {Multi-view learning for
  vision-and-language navigation}.
\newblock \emph{CoRR}, abs/2003.00857.

\bibitem[{Xiang et~al.(2020)Xiang, Wang, and Wang}]{DBLP:conf/emnlp/Xiang0W20}
Jiannan Xiang, Xin Wang, and William~Yang Wang. 2020.
\newblock \href {https://doi.org/10.18653/v1/2020.findings-emnlp.62} {Learning
  to stop: {A} simple yet effective approach to urban vision-language
  navigation}.
\newblock In \emph{Proceedings of the 2020 Conference on Empirical Methods in
  Natural Language Processing: Findings, {EMNLP} 2020, Online Event, 16-20
  November 2020}, pages 699--707. Association for Computational Linguistics.

\bibitem[{Zhu et~al.(2020{\natexlab{a}})Zhu, Zhu, Chang, and
  Liang}]{DBLP:conf/cvpr/Zhu0CL20}
Fengda Zhu, Yi~Zhu, Xiaojun Chang, and Xiaodan Liang. 2020{\natexlab{a}}.
\newblock \href {https://doi.org/10.1109/CVPR42600.2020.01003} {Vision-language
  navigation with self-supervised auxiliary reasoning tasks}.
\newblock In \emph{2020 {IEEE/CVF} Conference on Computer Vision and Pattern
  Recognition, {CVPR} 2020, Seattle, WA, USA, June 13-19, 2020}, pages
  10009--10019. {IEEE}.

\bibitem[{Zhu et~al.(2020{\natexlab{b}})Zhu, Wang, Fu, Yan, Narayana, Sone,
  Basu, and Wang}]{DBLP:journals/corr/abs-2007-00229}
Wanrong Zhu, Xin Wang, Tsu{-}Jui Fu, An~Yan, Pradyumna Narayana, Kazoo Sone,
  Sugato Basu, and William~Yang Wang. 2020{\natexlab{b}}.
\newblock \href {http://arxiv.org/abs/2007.00229} {Multimodal text style
  transfer for outdoor vision-and-language navigation}.
\newblock \emph{CoRR}, abs/2007.00229.

\bibitem[{Zhu et~al.(2020{\natexlab{c}})Zhu, Zhu, Zhan, Lin, Jiao, Chang, and
  Liang}]{DBLP:conf/cvpr/0004ZZLJCL20}
Yi~Zhu, Fengda Zhu, Zhaohuan Zhan, Bingqian Lin, Jianbin Jiao, Xiaojun Chang,
  and Xiaodan Liang. 2020{\natexlab{c}}.
\newblock \href {https://doi.org/10.1109/CVPR42600.2020.01074} {Vision-dialog
  navigation by exploring cross-modal memory}.
\newblock In \emph{2020 {IEEE/CVF} Conference on Computer Vision and Pattern
  Recognition, {CVPR} 2020, Seattle, WA, USA, June 13-19, 2020}, pages
  10727--10736. {IEEE}.

\end{thebibliography}
\bibliographystyle{acl_natbib}

\end{document}